\documentclass{article}
\usepackage{listings}
\usepackage{graphicx}
\usepackage{amssymb}

\begin{document}

\title{A Short Note on Modeling 2D Taut Ropes with Visibility Decompositions}
\markright{Modeling Taut Ropes}
\author{Adem Berat Dalk{\i}l{\i}\c{c}}

\maketitle

\begin{abstract}
The problem of modeling ropes arises in many applications, including providing haptic feedback to surgeons who are using surgical robots to realign the distal and proximal ends of split bones. Here, we consider a simplified, 2D variant of the haptic feedback estimation problem and discuss how visibility decompositions greatly simplify the problem. Then, we introduce an efficient, concise algorithm for modeling the dynamics of 2D ropes around polygonal obstacles in O($n$) time, where $n$ is the number of line segment obstacles.
\end{abstract}

\section{PROBLEM STATEMENT}
We start by providing a brief definition of our \textit{2D rope problem}. We define a constrained space as an open subset of $\mathbb{R}^2$, $C$, and the free space as $\mathbb{R}^2$ - $C$, or $F$. Two points in $\mathbb{R}^2$ are said to be \textit{visible} if the open line segment connecting them never lies in $C$. In this simplified version of the problem we define obstacles as open line segments made up of two points. The open line segment obstacles constitute $C$ entirely.

Consider a taut rope with two endpoints, $A$ and $B$, where $A$ and $B$ are initially visible to one another. Consider the trace created when $A$ is moved about such that it always remains in $F$. We parameterize this trace, such that $A(t)$ defines the positions of $A$ in $\mathbb{R}^2$ as a function of $t$. Because the trace is described as a set of connected line segments, the generation of such traces requires that for all $t$, $A(t-1)$ and $A(t+1)$ must both be visible to $A(t)$; if this condition was not true, then the rope would be passing through obstacle line segments, which is impossible. We define this requirement as the \textit{small update condition}. We also define the essential \textit{single-cut condition} in the next section.

If the shortest path is generated between $A$ and $B$ for each parameter $t$, then, excluding the cases where $A$ is visible to $B$, the path will contain at least one point of an obstacle line segment, which we call a \textit{wrapping point}. We can define the rope at parameter $t$, $R(t)$, as an ordered set of wrapping points contained by the trace between $A$ and $B$ at time $t$ if that trace were to be made taut; $R(t)$ is ordered by recency: the least recent wrapping point that was appended to $R(t)$ is the first element, and the most recently appended wrapping point is the last element.

To limit technicality, a very informal problem definition is presented. Given a trace, $A(t)$, which fulfills the small update and single-cut conditions, we wish to compute for all $t$ the taut rope made when the two ends of the trace are tightened without the locations of the endpoints of the trace actually changing.

\begin{figure}
    \centering
    \includegraphics[scale=0.5]{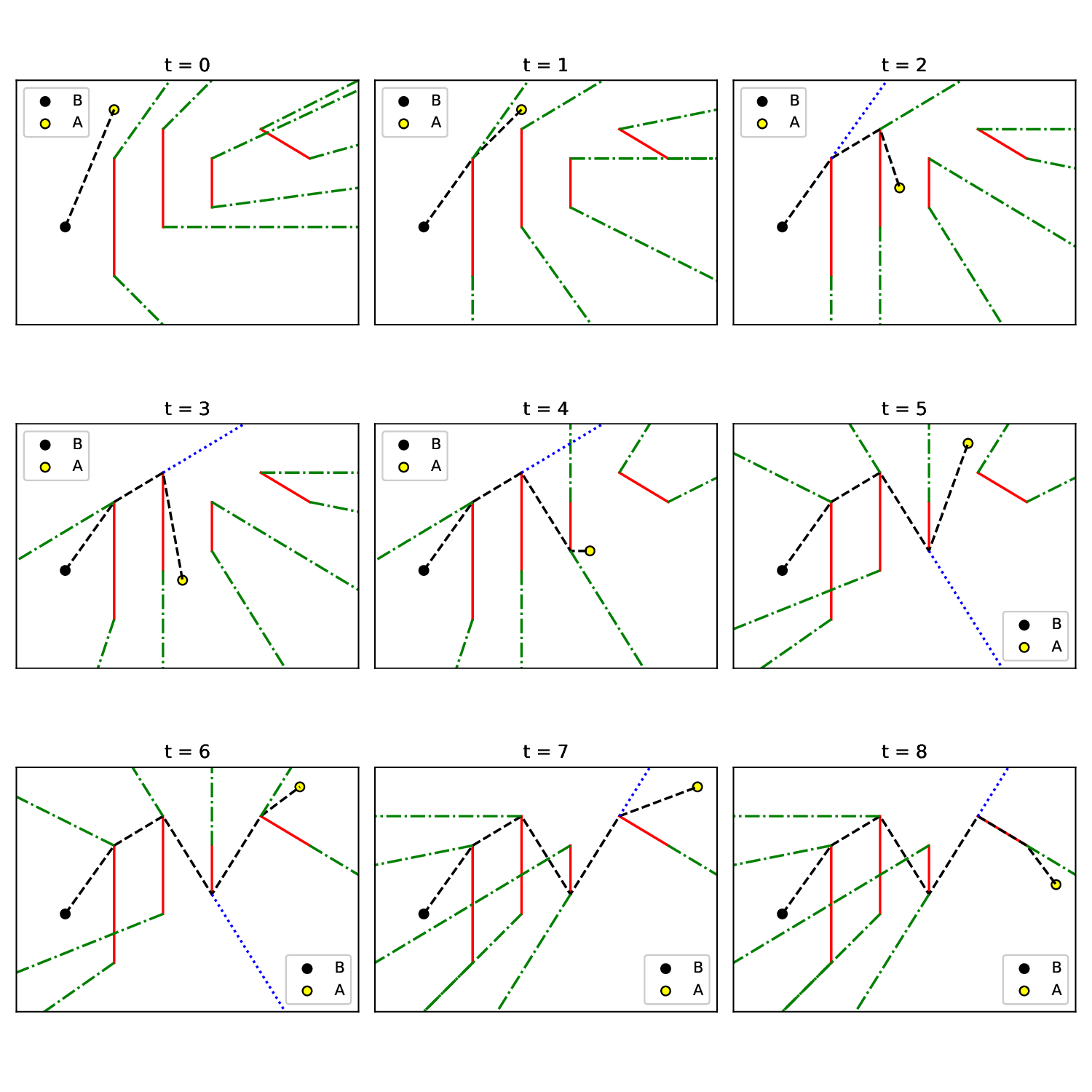}
    \caption{Visual depiction of the algorithm. $B$, the fixed end of the rope (filled circle) is shown, the rays of the GD that were computed until an intersection was found are shown (dash-dot lines), the unwrapping ray is shown (dotted), the rope is shown (dashed), and the obstacle line segments are shown (straight) for all $t$ of $A(t)$ (hollow circle).}
    \label{fig:enter-label}  

\end{figure}

\section{A SIMPLE SOLUTION}
The solution presented here requires that a partial visibility decomposition of $\mathbb{R}^2$ is computed. We call this pseudo-visibility decomposition a \textit{Gazan decomposition} (GD). The GD of $\mathbb{R}^2$ is the collection of all rays made by an observer point and all obstacle points; specifically, the rays begin from each obstacle point and extend away from the observer point infinitely with a slope equal to that of the line formed between the observer and obstacle point. The GD is useful because it provides a quantitative metric to determine when a new wrapping point should be added or removed from the rope. Specifically, when a ray of the GD is passed through by a movement of the trace, the source vertex of that ray must be appended to the rope. Because there are $n$ obstacle line segments, for every $t$, $n+1$ ray intersection tests must be performed between two entities: (1) the line segment defined by $A(t)$ and $A(t+1)$; (2) all rays of the GD with the most recent element of $R(t)$ as the observer point. To facilitate unwrapping, it is required to also include the most recently intersected ray as part of the rays to be checked for intersections, even if that ray is not part of the current GD. This is why there are $n+1$ intersection tests that must be performed instead of $n$. If it is shown that the normal from $A(t)$ to $A(t+1)$ passes through a ray of the GD of $R(t)$, then, the source vertex of that ray must be added to $R(t+1)$. However, if the ray is the same as the most recently passed through ray, then the source vertex of that ray must be removed from $R(t+1)$. The single-cut condition requires that, for all $t$, the norm created by the points $A(t)$ and $A(t+1)$ intersect at most one ray of not only the current GD, but also all the other GDs that are created when all other points are taken as observer points. This condition is important for preventing ambiguity when multiple rays are intersected simultaneously. Pythonic pseudocode for this algorithm is shown in listing 1.

\begin{lstlisting}[language=Python, caption={O($n$) Time Algorithm for 2D Rope Modeling}, basicstyle = \small]
obstacleSpace = [((Px1, Py1), (Px2, Py2)), . . ., 
                ((Px(f-1), Py(f-1)), (Pxf, Pyf))]
obstaclePoints = list(sum(obstacleSpace, ())) # flattens the list
rope = [B]
As = [(A1x, A1y), . . ., (Afx, Afy)]
A0 = As[0]
unwrappingRay = [(0,0), (0,0)]
for idx, A in enumerate(As[1:]):
    rays = []
    for obstaclePoint in obstaclePoints:
        if obstaclePoint == rope[-1]:
            continue
        ray = (obstaclePoint, (obstaclePoint[0] + Inf * 
                              (obstaclePoint[0] - rope[-1][0]),
                               obstaclePoint[1] + Inf * 
                              (obstaclePoint[1] - rope[-1][1])))
        rays.append(ray)
        if intersect(A0, A, ray[0], ray[1]):
            unwrappingRay = ray
            rope.append(ray[0])
            A0 = A
            break
        if not len(rope) < 2:
            if intersect(A0, A, unwrappingRay[0], unwrappingRay[1]):
                rope.pop()
                if len(rope) != 1:
                    unwrappingRay = (rope[-1], (rope[-1][0] + Inf * 
                                    (rope[-1][0] - rope[-2][0]),
                                    rope[-1][1] + Inf * 
                                    (rope[-1][1] - rope[-2][1])))
                else:
                    unwrappingRay = ((0,0),(0,0))
                A0 = A
                break
    A0 = A
\end{lstlisting}

\begin{figure}
    \centering
    \includegraphics[scale=0.5]{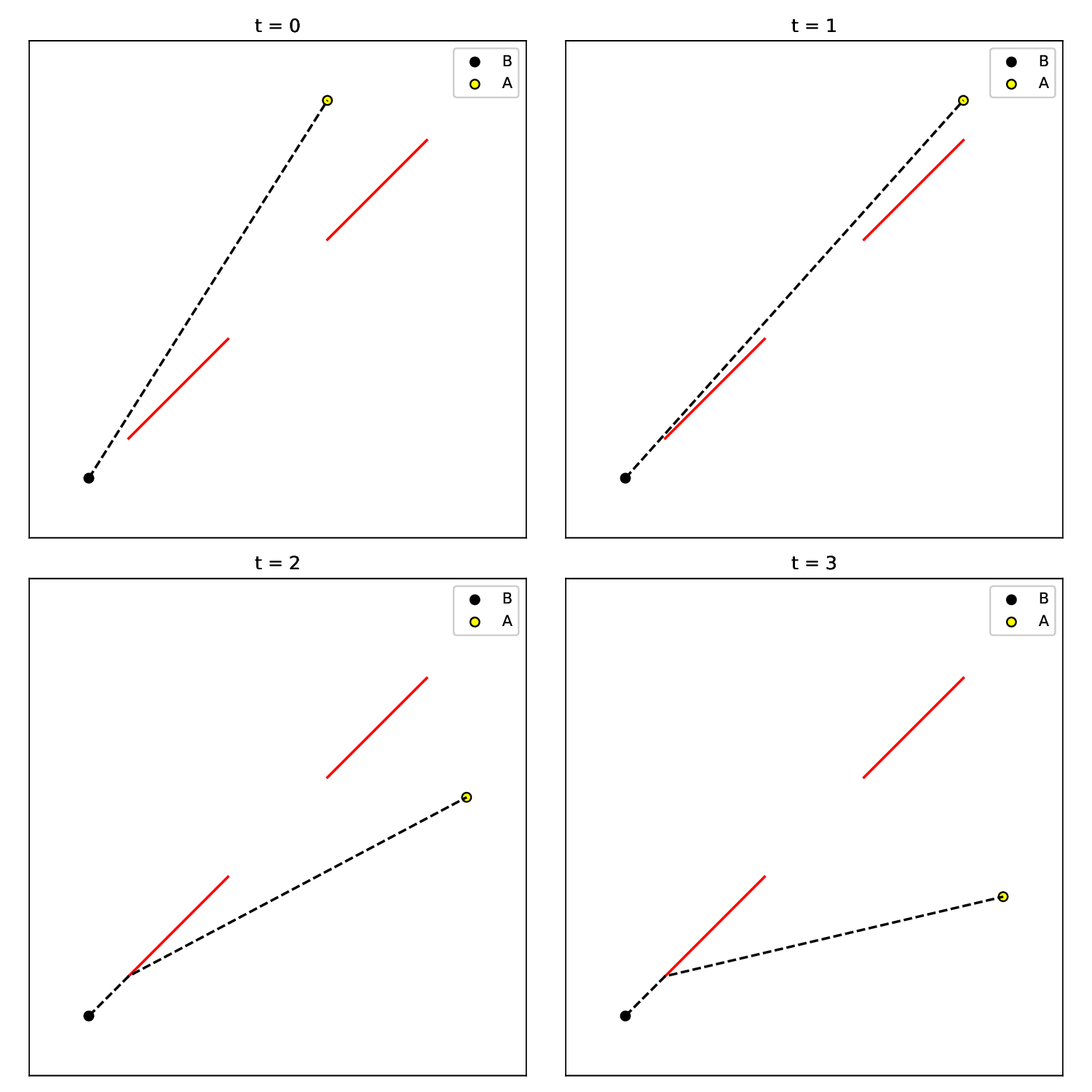}
    \caption{Failure case caused by colinearity. $B$, the fixed end of the rope (filled circle) is shown, the rope is shown (dashed), and the obstacle line segments are shown (straight) for all $t$ of $A(t)$ (hollow circle).}
    \label{fig:enter-label}
\end{figure}

\section{A SMALL LIMITATION}
If an obstacle space contains collinear line segments, then there is a possibility of ambiguity when determining which segments were intersected. For example, consider the case shown in figure 2. It is possible that the next movement would pass through 2 rays simultaneously, making the correct vertex to be chosen ambiguous. For this reason, this algorithm requires that there be no collinear obstacle line segments. However, this limitation can be resolved without increasing the time complexity: for all intersected rays, perform a linear search to identify the ray whose source is furthest from the most recent wrapping point of the rope; the furthest source should be the only appended vertex to the rope. Angles could also have been used as an alternative instead of rays when computing the GD, but rays are more convenient for reasons that the reader can discover on their own.

\section{AN OVERVIEW OF RELATED WORK FOR FURTHER READING}
The core of this algorithm is derived from the visibility decomposition technique, which was extensively explored in Boland's thesis \cite{boland2002polygon}. Visibility decompositions can be thought of as a type of space partitioning in which visibility to some set of vertices in the scene is the defining feature of each partitioned region \cite{gigus1990computing}. There have been many applications of visibility decompositions for solving 2D visibility query problems over different polygonal regions, especially for the visibility polygon problem. Bose et al. was probably the first to use the visibility decomposition to generate a data structure that facilitates O(log$n$) visibility query time for a query point inside or outside a single, simple polygon \cite{bose2002efficient}. The same decomposition had been described by Aronov et al. \cite{aronov2002visibility}, who used the decomposition to partition the interior of a polygon into cells; a point lying in any cell would have a unique visibility polygon. Zarei et al. \cite{zarei2008query} presented an extension of the visibility decomposition and provided a data structure for solving the query problem over a polygonal domain with holes. Inkulu and Kappor provided two improved algorithms that used the visibility decomposition to solve the query problem. Lu, provided alternative algorithms that incorporated aspects of the visibility decomposition technique \cite{lu2011point}. Shortly after, Baygi and Ghodsi developed another data structure that incorporated visibility decompositions to compute visibility polygons \cite{baygi2013space}. The previous techniques that utilize the visibility decomposition focused on creating data structures that facilitate O(log$n$) online query time by precomputing visibility decompositions. Here, however, our technique iteratively computes a lighter form of the visibility decomposition, the \textit{Gazan decomposition}, facilitating linear online time queries without any convoluted precomputation. Most importantly, we extend the visibility decomposition technique from traditional visibility querying to that of modeling taut ropes.
\bibliographystyle{plain}
\bibliography{main.bib}

\begin{thebibliography}{1}

\bibitem{aronov2002visibility}
Aronov, Guibas, Teichmann, and Zhang.
\newblock Visibility queries and maintenance in simple polygons.
\newblock {\em Discrete \& Computational Geometry}, 27:461--483, 2002.

\bibitem{baygi2013space}
Mostafa~Nouri Baygi and Mohammad Ghodsi.
\newblock Space/query-time tradeoff for computing the visibility polygon.
\newblock {\em Computational Geometry}, 46(3):371--381, 2013.

\bibitem{boland2002polygon}
Ralph~Patrick Boland.
\newblock {\em Polygon visibility decompositions with applications.}
\newblock University of Ottawa (Canada), 2002.

\bibitem{bose2002efficient}
Prosenjit Bose, Anna Lubiw, and J~Ian Munro.
\newblock Efficient visibility queries in simple polygons.
\newblock {\em Computational Geometry}, 23(3):313--335, 2002.

\bibitem{gigus1990computing}
Ziv Gigus and Jitendra Malik.
\newblock Computing the aspect graph for line drawings of polyhedral objects.
\newblock {\em IEEE Transactions on Pattern Analysis and Machine Intelligence}, 12(2):113--122, 1990.

\bibitem{lu2011point}
Lin Lu, Chenglei Yang, and Jiaye Wang.
\newblock Point visibility computing in polygons with holes.
\newblock {\em JOURNAL OF INFORMATION \&COMPUTATIONAL SCIENCE}, 8(16):4165--4173, 2011.

\bibitem{zarei2008query}
Alireza Zarei and Mohammad Ghodsi.
\newblock Query point visibility computation in polygons with holes.
\newblock {\em Computational Geometry}, 39(2):78--90, 2008.

\end{thebibliography}
\end{document}